\documentclass[11pt, letterpaper]{article}
\usepackage[top=0.85in,left=1in,footskip=0.75in,marginparwidth=2in]{geometry}

\usepackage{amsmath}
\usepackage{amssymb}
\usepackage{amsthm}
\usepackage{cases}
\usepackage{siunitx}
\usepackage{verbatim}
\usepackage{newclude}
\usepackage{bm}
\usepackage[sort&compress,comma,super]{natbib}

\usepackage[utf8]{inputenc}

\usepackage{nameref,hyperref}

\usepackage{microtype}
\DisableLigatures[f]{encoding = *, family = * }

\usepackage{changepage}

\usepackage[aboveskip=1pt,labelfont=bf,labelsep=space,singlelinecheck=off]{caption}

\makeatletter
\renewcommand{\@biblabel}[1]{\quad#1.}
\makeatother

\usepackage{lastpage,fancyhdr,graphicx}
\usepackage{epstopdf}

\usepackage{color}

\definecolor{Gray}{gray}{.25}

\usepackage{graphicx}

\usepackage{sidecap}
\renewcommand{\figurename}{Fig.}
\usepackage{wrapfig}
\usepackage[pscoord]{eso-pic}
\usepackage[fulladjust]{marginnote}
\usepackage{siunitx}
\reversemarginpar

\usepackage{booktabs}
\usepackage{multirow}

\begin{document}
\vspace*{0.35in}
\definecolor{bured}{rgb}{0.8, 0.0, 0.0}

\begin{flushleft}
{\Large
\textbf\newline{Towards free-response paradigm: a theory on decision-making in spiking neural networks}
}
\newline
\\
Zhichao Zhu\textsuperscript{1,2}, 
Yang Qi\textsuperscript{1,2,3}, 
Wenlian Lu\textsuperscript{5,6,7,8},
Zhigang Wang\textsuperscript{9},
Lu Cao\textsuperscript{9},
Jianfeng Feng\textsuperscript{1,2,3,4,*}
\\
\bigskip
\it{1} Institute of Science and Technology for Brain-Inspired Intelligence, Fudan University, Shanghai, China
\\
\it{2} Key Laboratory of Computational Neuroscience and Brain-Inspired Intelligence (Fudan University), Ministry of Education, China
\\
\it{3} MOE Frontiers Center for Brain Science, Fudan University, Shanghai, China
\\
\it{4} Zhangjiang Fudan International Innovation Center, Shanghai, China
\\
\it{5} School of Mathematical Sciences, Fudan University, Shanghai, China
\\
\it{6} Shanghai Center for Mathematical Sciences, Shanghai, China
\\
\it{7} Shanghai Key Laboratory for Contemporary Applied Mathematics, Shanghai, China
\\
\it{8} Key Laboratory of Mathematics for Nonlinear Science, Shanghai, China
\\
\it{9} Intel Labs China, Beijing, China
\bigskip

\end{flushleft}
\section*{Abstract}
The energy-efficient and brain-like information processing abilities of Spiking Neural Networks (SNNs) have attracted considerable attention, establishing them as a crucial element of brain-inspired computing.
One prevalent challenge encountered by SNNs is the trade-off between inference speed and accuracy, which requires sufficient time to achieve the desired level of performance. 
Drawing inspiration from animal behavior experiments that demonstrate a connection between decision-making reaction times, task complexity, and confidence levels, this study seeks to apply these insights to SNNs.
The focus is on understanding how SNNs make inferences, with a particular emphasis on untangling the interplay between signal and noise in decision-making processes. 
The proposed theoretical framework introduces a new optimization objective for SNN training, highlighting the importance of not only the accuracy of decisions but also the development of predictive confidence through learning from past experiences. 
Experimental results demonstrate that SNNs trained according to this framework exhibit improved confidence expression, leading to better decision-making outcomes. 
In addition, a strategy is introduced for efficient decision-making during inference, which allows SNNs to complete tasks more quickly and can use stopping times as indicators of decision confidence. 
By integrating neuroscience insights with neuromorphic computing, this study opens up new possibilities to explore the capabilities of SNNs and advance their application in complex decision-making scenarios.
\section{Introduction}
Spiking neural networks (SNNs) are considered the third generation of neural networks and are believed to hold promise in the implementation of more complex cognitive functions in a manner that closely mimics the processing principles of the brain cortex~\cite{Maass1997, Roy2019}. 
Despite the significant advances made by Artificial neural networks (ANNs, the second generation of neural networks) in various fields~\cite{LeCun2015, Wang2023}, it is still unclear how to maximize the potential of SNNs~\cite{Pfeiffer2018, Neftci2019, Li2023}.
Although many investigations have effectively conducted training on SNNs \cite{Wu2018, shrestha2018slayer, Bellec2020, Rueckauer2017, Wu2022, deng2022temporal}, a common issue encountered by these networks is the inherent balance between speed and accuracy~\cite{Deng2020, Dong2024} that achieving superior performance often requires longer simulation time.

Instead of concentrating on decreasing the latency of SNN for the desired accuracy, we adopt the principles of human cognitive decision-making that consider latency as an indicator of decision confidence.
The core aspect of perceptual decision-making involves the process of accumulating evidence, which is a method used to address uncertainties in the environment by integrating information over time~\cite{Mazurek2003, Churchland2008, Palmer2005, Gold2007}.
This integration process embodies the speed-accuracy trade-off~\cite{Gold2007, Bogacz2010}: the likelihood of a correct decision improves with the gradual aggregation of evidence, yet this collection of information inherently demands time. 
The reaction time, which captures the duration from stimulus presentation to response, varies with task difficulty~\cite{Kepecs2008, Kiani2014}, and is significantly shorter for correct decisions compared to incorrect ones~\cite{Vickers1982, Deco2009, Kiani2014}, indicating its potential as an index of confidence in decisions.

Random drift-diffusion models (DDMs) are widely used to understand the fundamental aspects of psychophysical tasks~\cite{Gold2007, Bogacz2006, Ratcliff2004, Ratcliff2008}. 
Within the domain of DDMs, the decision-making process involves accumulating evidence, which is depicted as the integral of a stochastic process over time until a specific decision threshold is reached. 
This conceptual framework not only corresponds to experimental findings, but also suggests that the brain employs probabilistic reasoning through the sampling of neural activity~\cite{hoyer2002interpreting, savin2014spatio, Haefner2016}. Consequently, the precision of decisions is affected by both the strength of the evidence (captured by the mean) and the variability of that evidence (captured by the covariance)~\cite{Averbeck2006, AzeredodaSilveira2021, Panzeri2022}.

In this study, we explore the SNN inference processes to establish a theoretical link between decision-making and decision confidence, in line with the principle used in DDMs.
This theory highlights the role of mean and covariance in affecting the necessary latency to make a decision, which is comparable to reaction times in neuroscience. 
As a result, it changes the decision-making process in SNNs from the traditional interrogation paradigm, where decisions are made within a set time frame regardless of input, to the free-response paradigm where SNNs dynamically stop the inference process based on the confidence level of the current input.

To demonstrate our theory, we introduce a fidelity-entropy loss function aimed at guiding SNNs not just towards making accurate decisions but also towards enhancing their confidence levels based on the decision outcomes. 
Through experiments carried out at various levels of difficulty, we show that this loss function enhances confidence expression and speeds decision-making.
Moreover, our theory introduces a successful stopping strategy that improves the speed of model inference by allowing flexible termination, with the stopping time indicating the level of confidence. 
The efficiency of this stopping approach in various SNN training frameworks highlights the importance of temporal dynamics in SNNs, which provides a fresh outlook on latency in SNNs.
\section{Results}
\subsection{Theoretical framework for decision-making in spiking neural networks}
\label{sec:1-theory}
\begin{figure}
  \begin{center}
  \includegraphics[width=\linewidth]{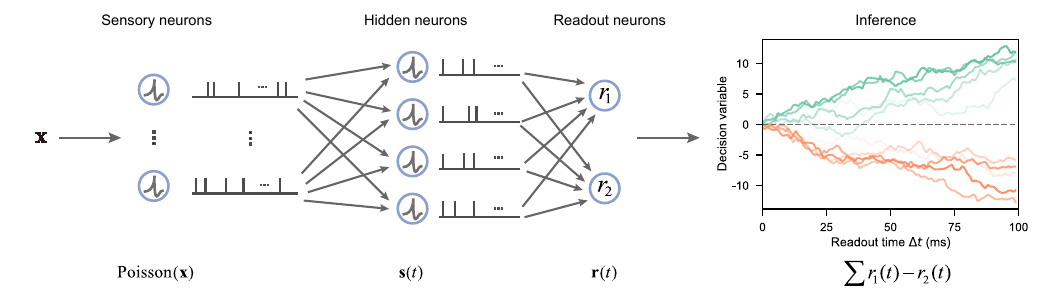}
  \end{center}
  \caption{\textbf{The drift-diffusion model framework applied to SNN decision-making.} Sensory neurons encode stimulus \(\mathbf{x}\) (e.g., an image) into corresponding spike trains via a Poisson process (\(\text{Poisson}(\mathbf{x})\)). 
  These trains serve as input to a downstream network, which processes the information through successive hidden layers. 
  The final layer of the network employs a linear readout to transform spiking activity \(\mathbf{s}(t)\) at time \(t\) into evidence \(\mathbf{r}(t)\) for decision-making. 
  The decision variable integrates and compares evidence over time (\(\sum_{t=1}^{T} (r_1(t) - r_2(t))\)), emulating evidence accumulation in the drift-diffusion model, to determine the point of commitment to the decision or the need for additional evidence collection.} \label{fig:1}
  \end{figure}

We begin with an overview of the classification task pipeline in SNNs, as illustrated in \figurename~\ref{fig:1}.
The initial encoding of an external stimulus \(\mathbf{x}\) (e.g. an image) into neural spike activity is facilitated by sensory neurons through rate coding. 
In this process, the intensity of each image pixel is proportionally reflected in the firing rate of the corresponding neuron. 
These spike trains are then the input to the SNN. 
Typically, the decision-making process uses the activity of neurons of the network output layer, denoted by \(\mathbf{s}(t)\) at time \(t\). 
A linear decoder, represented by a weight matrix \(W\), continuously maps the spiking activity \(\mathbf{s}(t)\) to a decision readout \(\mathbf{r}(t)\), given by:
\begin{equation}
  \mathbf{r}(t) = W\mathbf{s}(t).
  \label{eq:instantaneous_readout}
\end{equation}
The instantaneous readout value \(\mathbf{r}(t)\) is inherently subject to randomness due to the presence of input noise originating from Poisson-distributed spike trains and internal network variations. 
These variations include a randomly distributed initial membrane potential of neurons, a rough time resolution to model the behavior of neurons to integrate and fire, and a restricted precision of synaptic weights.
Therefore, the instantaneous readout \(\mathbf{r}(t)\) can be modeled as sampling from a distribution characterized by the likelihood function \(p(\mathbf{r}|\mathbf{x}, \theta)\)~\cite{savin2014spatio, Zhang2023}, where \(\theta\) represents the SNN parameters.
Reliable inference is dependent on a sufficiently extended readout window, which helps to strike a balance between performance and efficiency in SNNs~\cite{Deng2020}. 
Namely, rather than relying on instantaneous readout, we accumulate evidence over time to construct a cumulative readout \(\mathbf{d}\), which is defined as:
\begin{equation}
  \mathbf{d} = \sum_{t=1}^T \mathbf{r}(t).
  \label{eq:cumulative_readout}
\end{equation}
To ensure clarity, we denote the readout interval from the start of the readout process at time 0 to the current time \(T\) as \(\Delta t\).
Assuming that each \(\mathbf{r}(t)\) is drawn from a Gaussian distribution with mean \(\mathbf{\mu}\) and covariance \(\mathbf{\Sigma}\) independently, then \(\mathbf{d} \sim \mathcal{N}(\mathbf{\mu}\Delta t, \mathbf{\Sigma}\Delta t)\).
As \(\Delta t\) expands, the readout evidence progressively aligns with \(\mu \Delta t\), culminating in high-fidelity evidence conducive to robust decision-making.

Consider a binary decision-making task wherein the process is distilled to evaluate the sign of the decision variable (DV), defined as \(d_i - d_j\) for the two corresponding entries in \(\mathbf{d}\).
In the context of noise, potential DV trajectories over time (as depicted in \figurename~\ref{fig:1}, the rightmost panel) may exhibit initial fluctuations but will stabilize given a sufficiently long readout interval. 
If we model the probability of each category by the softmax function:
\begin{equation}
  P(i|\mathbf{d}) = \frac{\exp d_i}{\sum_j \exp d_j}.
  \label{eq:softmax_func}
\end{equation}
Then the log-likelihood ratio between category \(i\) and \(j\) is:
\begin{equation}
  \Lambda = \ln \frac{P(i|\mathbf{d})}{P(j|\mathbf{d})} = d_i - d_j.
\end{equation}
Consequently, a greater difference between \(d_i\) and \(d_j\) suggests a higher probability of winning by selecting category \(i\) over \(j\).
Moreover, when the readout time \(\Delta t\) is long enough, the decision metric approaches the measurement of \((\mu_i-\mu_j)\Delta t\). 
This guarantees that the model favors the category with the highest probability, in accordance with the principle of maximum likelihood estimation.
However, \(\Delta t\) is often limited due to the importance of quick responses. Therefore, decision-making in SNNs aims to answer two main questions: Which category should be selected as the most suitable choice? And is the chosen option truly the best one based on the available evidence?

The first question, which identifies the best category, focuses on the index of the maximal entry in \(\mathbb{d}\) (the sign of the DV). 
Notably, since the variance of the DV's magnitude grows at a slower rate than its mean, there comes a point when the sign of the DV stabilizes. 
This moment marks the convergence time of the inference process, where a shorter convergence time implies a faster commitment to a prediction by the SNNs.
The second inquiry pertains to the level of confidence in that choice, which is determined by the distance from the decision boundary (\(d_i = d_j\)). Rather than providing a direct response, we approach it by assessing the likelihood of an SNN accurately recognizing the category with the highest average rate. 
Assuming that the category \(i\) exhibits the highest mean rate, the probability that the network will make the best decision (selecting \(i\)) is equivalent to computing the probability \(P(d_i - d_j > 0)\), which signifies the confidence of the model in its current decision.
When we presuppose that the readouts \(\mathbf{r}\)  are drawn from a Gaussian distribution with mean \(\mathbf{\mu}\) and covariance \(\mathbf{\Sigma}\), we obtain an analytical expression for this probability:
\begin{equation}
  P(d_i - d_j > 0) = \frac{1}{2} \text{erfc} (- \frac{(\mu_i - \mu_j)\sqrt{\Delta t}}{\sqrt{2(\sigma_i^2 + \sigma_j^2 - 2\sigma_i \sigma_j \rho_{ij})}}),
  \label{eq: fidelity}
\end{equation}
where \(\sigma_i, \sigma_j\) represent the standard deviations of the respective categories, \(\rho_{ij}\) is the correlation coefficient between categories \(i\) and \(j\) and \(\rm erfc\) denotes the complementary error function.

\begin{figure}
  \begin{center}
      \includegraphics[width=\linewidth]{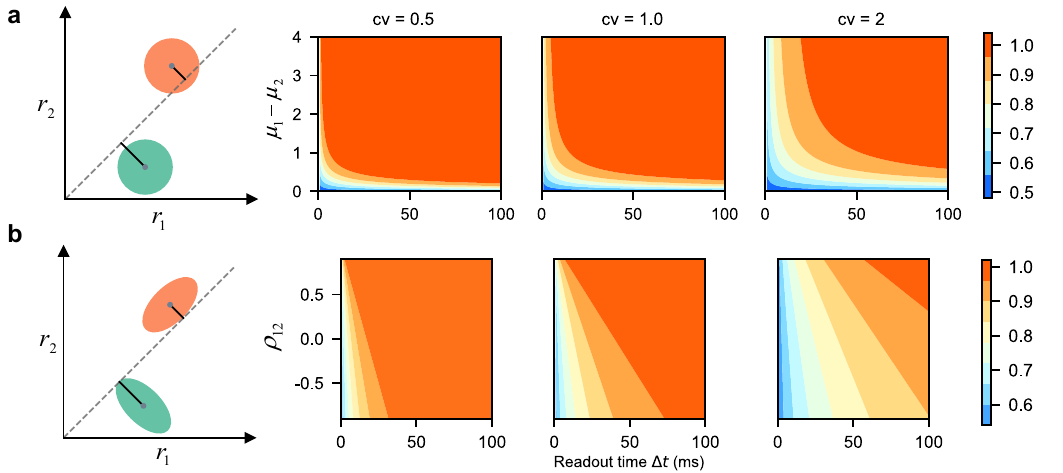}
  \end{center}
  \caption{\textbf{Influence of readout dynamics on model confidence.} \textbf{a}, Model confidence, defined as \(P(d_i - d_j > 0)\) is quantified by the distance between the readout means (\(\mu_i, \mu_j\)) and the classification boundary, incorporating variance in the assessment.
  The analysis, which fixes \(\mu_j = \sigma_j = 1, \rho_{ij} =0\), explores the evolution of confidence in readout time \(\Delta t\) as \(\mu_i\) varies and \(\sigma_i\) adjusts according to the coefficient of variation (CV).
  \textbf{b}, The effect of correlation between readouts on decision confidence is examined under a consistent mean disparity and modified CVs. 
  How confidence \(P(d_i - d_j > 0)\) changes under different conditions in \textbf{a} and \textbf{b} is illustrated by color maps.}
  \label{fig:2}
\end{figure}

The key determinant influencing the decision is the disparity in means \(\mu_i - \mu_j\), as this is the only factor dictating the direction of the confidence boost.
Therefore, the confidence of the model in the suitability of category \(i\) will increase only when \(\mu_i\) has the highest mean as the readout time \(\Delta t\) increases. 
This criterion ensures that, given enough readout time, the system will consistently prefer the option with the highest mean value. 
Furthermore, a greater difference in means can speed up confidence within shorter readout periods, as illustrated in the upper panels of \figurename\ref{fig:2}.
On the other hand, in a scenario with constant parameters, a higher variance can hinder this progression, as illustrated in the same graphical representations where \(\rho_{ij}\) is maintained at zero. 
Nonetheless, this impact can be influenced by the covariance configuration (see \figurename\ref{fig:2}, lower graphs). 
When there is a positive correlation between entry \(i\) over \(j\), indicating alignment of the principal axis of the covariance matrix \(\Sigma\) with the decision boundary, it reduces the denominator in Eq. (\ref{eq: fidelity}), thus speeding up the growth of confidence. 
Conversely, negative correlations will produce the opposite outcome, slowing the increase in confidence.

From the preceding discussion, we can distill two primary training objectives for SNNs. 
The first objective focuses on ensuring that the correct category possesses the highest mean rate, since the ratio \(\mathbf{d}/\Delta t\) serves as an unbiased estimator of \(\mathbf{\mu}\), guiding the selection of the largest entry for decision-making. 
The second objective is to enhance the confidence of the model in accurately predicted samples while reducing it for incorrect predictions.
However, in the context of multi-class decision-making, the second objective confidence introduces computational challenges. 
Specifically, confidence involves calculating the probability \(q_i = \int p(\mathbf{d} \mathbf{x}, \theta)\mathbf{1}_i(\mathbf{d})d\mathbf{d}\), where the indicator function \(\mathbf{1}_i(\mathbf{d})\) equals one if \(d_i\) is the maximum among all entries and zero otherwise.
Due to the lack of a straightforward analytical solution for this probability and the computational cost associated with sampling-based estimations in high-dimensional spaces, our theoretical framework adopts a more pragmatic approach by focusing on the confidence comparison between the largest entry and the others pairwise, based on the mean vector \(\mathbf{\mu}\).
The simplification is justified because the confidence expression, as explained in Equation \ref{eq: fidelity}, is primarily impacted by the mean vector \(\mathbf{\mu}\) when \(\Delta t\) is large. 
Therefore, a high probability \(P(d_i - d_j > 0)\), indicating a clear preference for the highest value at the index \(i\) over the second highest at index \(j\) in \(\mathbf{\mu}\), implies inherently that \(P(d_i - d_k > 0)\) is also substantially high for all \(k\neq i\).
Essentially, if the model confidently favors one element over its closest competitor, it is likely to be equally confident about all other elements.
This understanding leads to the formulation of a novel loss function for SNN optimization
\begin{equation}
  L = \underbrace{\left(-\mu_{\rm label} + \sum_k \exp \mu_k \right)}_{\text{cross-entropy loss}} + \underbrace{f(\text{label}, i)\left(-\sum_{k\neq i} w_k \left(P_{ik} \log P_{ik} + (1-P_{ik})\log (1-P_{ik})\right)\right)}_{\text{fidelity-entropy loss}},
  \label{eq: fidelity_loss}
\end{equation}
where \(\mu_{\rm label}\) is the mean of the target entry, \(P_{ik}\) is the confidence measurement [Eq. (\ref{eq: fidelity})] between the top entry \(i\) and others with \(w_k\) as a nonnegative weight summing to 1, and \(f(\text{label}, i)\) is an indicator function that aligns optimization with prediction accuracy, producing 1 for correct predictions and -1 for incorrect predictions. 
The cross-entropy component is focused on enhancing the mean of target entry, in line with conventional training methods for ANN and SNN. 
On the other hand, the fidelity entropy component is intended to calibrate the model's confidence level based on the accuracy of predictions. 
In this context, \(\Delta t\) functions as a hyperparameter that balances the optimization objective between the numerator and the denominator of Eq. (\ref{eq: fidelity}).
In particular, when \(\Delta t\) is large, confidence is predominantly influenced by the numerator, \(\mu_i - \mu_j\). 
This underscores the importance of the fidelity entropy between the most confident prediction and the second most confident after exceeding a certain threshold, as it automatically reflects the condition met when comparing the entry \(i\) with the remaining entries. 
Consequently, only the fidelity entropy related to the two highest entries \(i\) and \(j\) is relevant. 
Therefore, \(w_j\) needs to have a significant value to have a greater impact on fidelity-entropy loss. 

Furthermore, as described in Equation \ref{eq: fidelity}, the time taken for the readout process \(\Delta t\) to reach a decision can vary depending on the inputs, which have unique effects on the distribution of the readout. 
By setting a decision threshold \(\Theta\), similar to the thresholds used in DDMs, a model can provide a clear prediction when \(P(d_i - d_j > 0) > \Theta\).
This condition delineates the minimal readout time as:
\begin{equation}
    \Delta t > (\text{erfc}^{-1}(2\Theta))^2 * \frac{2(\sigma_i^2 + \sigma_j^2 - 2\sigma_i\sigma_j\rho_{ij})}{(\mu_i -\mu_j)^2}.
    \label{eq:minimal_time}
\end{equation}
This equation explicitly indicates that \(\Delta t\) the minimum duration required to exceed a specified confidence threshold inherently fluctuates in response to the readout mean and covariance, which is influenced by the input \(\mathbf{x}\).
Consequently, this introduces a paradigm shift from an interrogation paradigm, in which models make decisions with a fixed readout time window regardless of various inputs, to a free-response paradigm that uses confidence levels to determine when to make a decision, aligning with the principles seen in DDMs~\cite{Bogacz2006, Gold2007, Ratcliff2008}.

To validate our hypothesis, we utilize the moment neural network (MNN) as a substitute for SNN training. 
The MNN, referenced in seminal works~\cite{Feng2006, Lu2010, Qi2022}, is chosen for its mathematical rigor and its ability to accurately replicate the statistical properties and nonlinear dynamics inherent in the correlated fluctuations observed in spiking neurons. 
This ability enables direct optimization of the readout distribution, a crucial component of our methodology.
A brief introduction to MNN is provided in the Methods section. 
Our forthcoming analysis will emphasize the improved confidence representation of MNNs trained with the proposed fidelity-entropy loss, compared to those trained without it. 
Subsequently, we will proceed to an experimental validation phase where SNNs, constructed using the trained MNN parameters, will be scrutinized. 
This stage is intended to underscore the effectiveness of our theoretical framework. 
Furthermore, we will investigate different policies that aim to identify convergence times. 
The implementation of these policies will demonstrate the practical utility of the free-response paradigm, effectively combining theoretical insights with practical advances in SNNs.

\subsection{Empirical validation and superiority of introducing fidelity-entropy loss in moment neural network training}
We conducted experiments using three well-established datasets: MNIST~\cite{Lecun1998}, Fashion-MNIST (F-MNIST)~\cite{xiao2017fashion} and CIFAR10~\cite{krizhevsky2009learning}.
For MNIST and F-MNIST, our MNN implementation featured a single hidden layer, while for CIFAR10, we employed a more complex model with three hidden layers. 
The architectural configurations are not intended to show the effectiveness of training, but to conduct a comprehensive assessment of how well the models can demonstrate confidence when trained with and without the fidelity-entropy loss, referred to as \textit{WithFidelity} and \textit{MeanOnly}, under varying levels of task complexity, since models can achieve high accuracy in MNIST, their performance on the more demanding CIFAR10 dataset will be notably limited under this setup.
Detailed descriptions of the experiments are provided in the Methods.

\begin{figure}
    \centering
    \includegraphics[width=\linewidth]{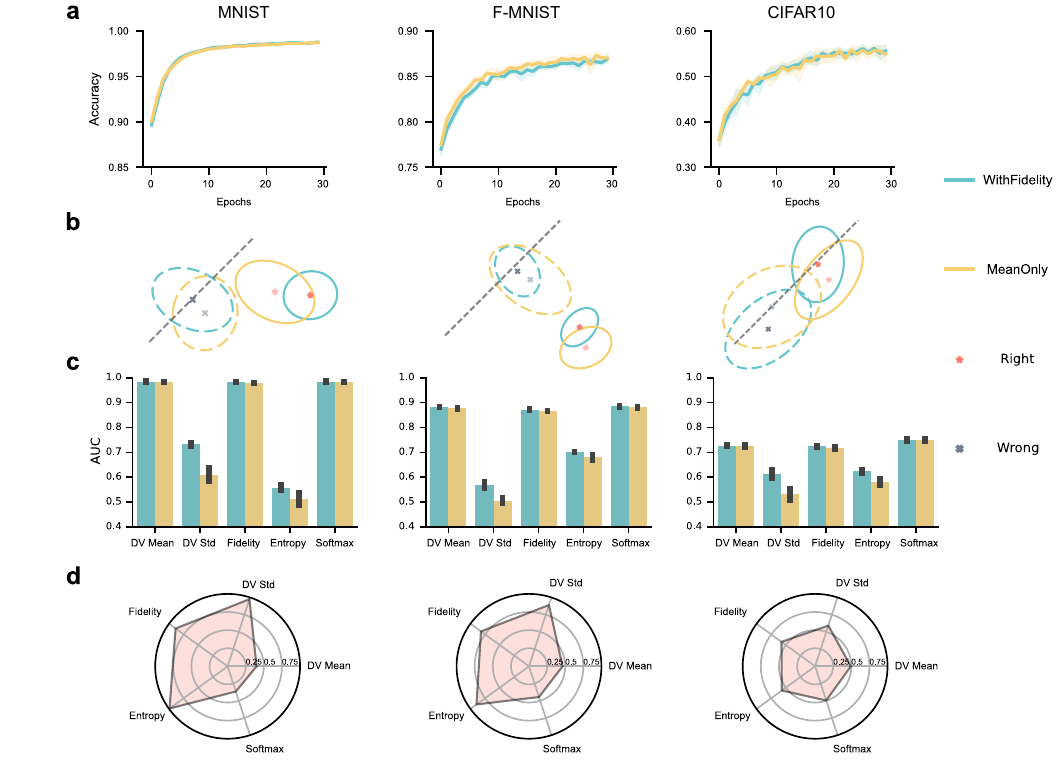}
    \caption{Efficacy of fidelity-entropy loss in Training MNNs. 
    \textbf{a}, Training accuracy over epochs for MNNs, comparing models trained with and without fidelity-entropy loss. The shaded regions denote the 95\% confidence intervals.
    \textbf{b}, Case study examples showing the influence of fidelity-entropy loss on model outputs. Correct predictions (red stars) and incorrect predictions (black crosses) are plotted based on the top two readout means, with their covariances represented by ellipses. Solid lines indicate covariances for correct cases, while dashed lines correspond to incorrect ones. The decision boundary is depicted as a dashed line.
    \textbf{c}, The models' ability to differentiate between correct and incorrect predictions is evaluated by the area under the receiver operating characteristic curve (AUROC), using various output metrics.
    \textbf{d}, Radial charts illustrating the proportion of samples where the application of fidelity-entropy loss enhances performance across selected metrics.}
    \label{fig:3}
\end{figure}

In our initial analysis, we focused on comparing the accuracy of the test set throughout the training period.
Every dataset underwent five experimental runs, each run comprising 30 epochs.
Models trained with fidelity-entropy loss (\textit{WithFidelity}) and those trained without it (\textit{MeanOnly}) showed similar levels of accuracy throughout training epochs, as illustrated in \figurename~\ref{fig:3}\textbf{a} (refer also to Table \ref{tab:S1}).
This finding is consistent with our theoretical framework, which indicates that the accuracy of decision-making is mainly dependent on the average rates, optimized by the cross-entropy component in Eq.~(\ref{eq: fidelity_loss}). 
It is crucial to note that integrating the fidelity-entropy loss does not result in a compromise in performance, highlighting its harmony with preserving model performance. 
As the fidelity component aims to improve the model's prediction confidence by boosting its capacity to precisely determine the probability of accurate versus inaccurate predictions.

The inclusion of fidelity-entropy loss does not ensure a consistent improvement in confidence levels across all samples. 
The reaction of the model to different inputs can lead to three possible outcomes (see \figurename~\ref{fig:3}\textbf{b}). 
In the most favorable scenario (on the left), \textit{WithFidelity} effectively pushes correct predictions (highlighted by a red star) away from the decision boundary (represented by a dashed line), reducing covariance and aligning the major axis of the covariance towards the boundary (illustrated by a solid line ellipse), a behavior not observed in (\textit{MeanOnly}) models. 
Conversely, incorrectly classified samples (depicted by a black cross) are pushed towards the boundary with an increased covariance (dashed line ellipse). 
In intermediate situations (middle panel), the model may optimize either the mean or the covariance, but not both, making it challenging to consistently boost confidence levels. In the worst-case scenario (right panel), fidelity loss might misguide optimizations, eroding confidence in correct predictions, while spuriously inflating confidence in incorrect ones. 

To evaluate the impact of fidelity-entropy loss on model performance, we utilized Area Under the Receiver Operating Characteristic (AUROC) analysis to evaluate the models' ability to distinguish between correct and incorrect predictions~\cite{Fawcett2006}.
Our analysis focused on five key metrics: DV Mean [the difference of the two top entries in mean, the numerator in Eq.~(\ref{eq: fidelity})], DV Std [the denominator in Eq. (\ref{eq: fidelity})], Fidelity [a combination of DV Mean and DV Std, as formulated in Eq. (\ref{eq: fidelity})], Entropy [based on output covariance, as formulated in Eq.~(\ref{eq:gaussian_entropy})], and Softmax (the highest softmax score of the readout mean). 
As demonstrated in \figurename~\ref{fig:3}\textbf{c}, metrics that incorporate the mean (DV Mean, Fidelity, Softmax) are the most effective in distinguishing decision outcomes, with superior AUROC scores indicating better performance. 
Improvement in these mean-related metrics depends on the overall accuracy of the model, as higher accuracy correlates with higher AUROC scores. 
Both the models \textit{WithFidelity} and \textit{MeanOnly} prioritize enhancing the mean, resulting in similar AUROC scores for these metrics, albeit with slightly better results from the \textit{WithFidelity} models. 
The most notable advancements are evident in the DV Std and Entropy metrics. 
Models trained using \textit{WithFidelity} exhibit significantly higher AUROC values in these aspects compared to the \textit{MeanOnly} models, which hover around the baseline of 0.5. 
Therefore, the fidelity-entropy loss plays a crucial role in optimizing the covariance structure of the model.

We also assessed the probability that the fidelity-entropy loss can boost confidence expression. 
For instances that are correctly classified, it is expected that metrics such as the DV mean, Fidelity, and Softmax will show an increase, whereas the DV Std and Entropy are likely to decrease. 
Conversely, for misclassified instances, the opposite trend is expected.
A probability greater than 0.5 indicates a positive impact.
As illustrated in \figurename~\ref{fig:3}\textbf{d}, the DV std and Entropy metrics consistently exhibit a noticeable improvement across all datasets, highlighting the role of the fidelity-entropy loss in optimizing the output covariance. 
Conversely, the improvements in mean DV and Softmax display a slight negative trend. However, due to the substantial enhancement in DV Std, there is also a successful enhancement in the Fidelity metric. 
The degree of these improvements is also linked to the model's accuracy, with notable advantages observed in high-performing models such as those trained on MNIST and F-MNIST, in comparison to CIFAR10. 
This discrepancy arises because the optimization direction in the fidelity-entropy loss is contingent on the prediction outcome. 
In challenging tasks, the input's correctness may vary between training iterations, resulting in a nearly negligible optimization effect due to frequent changes in direction.
 
In summary, according to the MNN viewpoint, the fidelity-entropy loss does not undermine the accuracy of classification. 
Although it may not increase the confidence expression for individual samples, it generally improves the model's output covariances. 
These enhancements are strongly associated with the model's initial accuracy, indicating that the influence of the fidelity-entropy loss differs depending on the complexity of the task.
In the following section, we will investigate how the theoretical improvements in MNNs translate into practical benefits in SNN performance.

\subsection{Fidelity-entropy loss leads to more consistent predictions and shorter convergence times in SNNs}
Leveraging the intrinsic link between MNN and SNN, we seamlessly transition to SNNs by reconstructing them using parameters derived from the MNN models discussed previously. 
Specifically, we utilize the parameters from the two MNN models depicted in \figurename \textbf{b} as a basis for constructing the corresponding SNN models, aiming to elucidate the enhancements fidelity-entropy loss brings to SNNs. 
To thoroughly assess performance, we carried out 100 trials for every sample in the test set, where each test lasted 100 milliseconds (equivalent to 100 time steps).
Detailed experimental settings are outlined in the Methods.

\begin{figure}
    \centering
    \includegraphics[width=\linewidth]{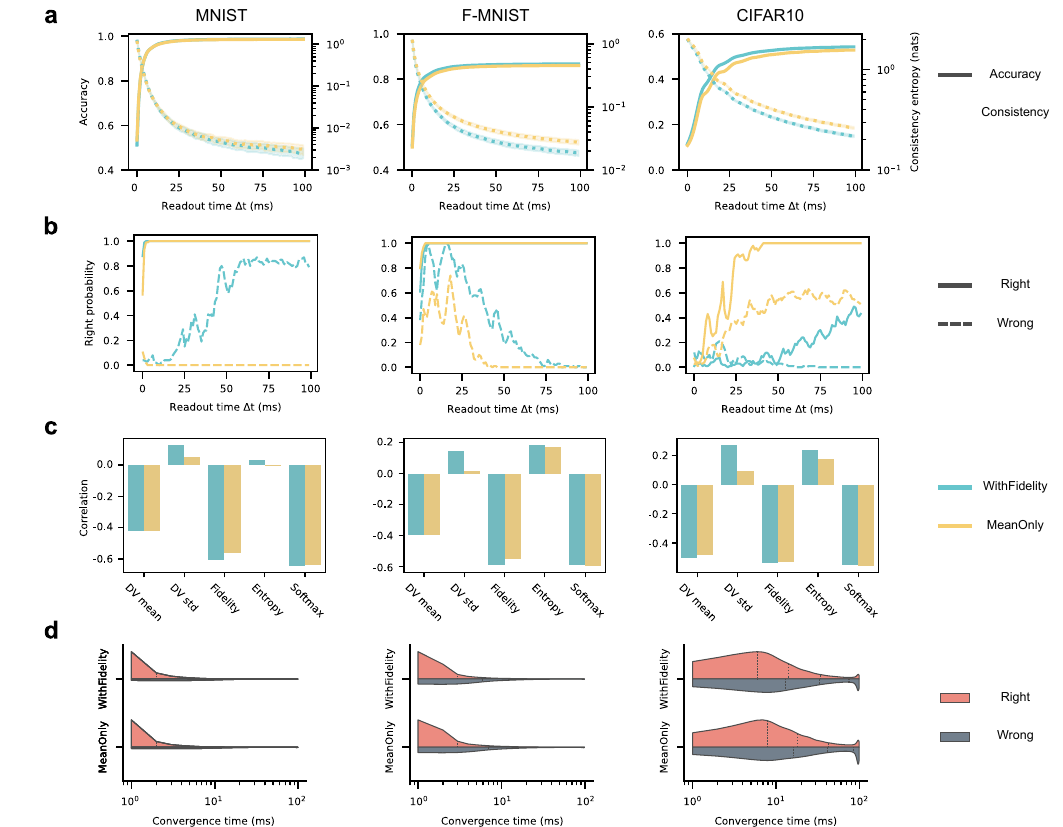}
    \caption{\textbf{Fidelity-entropy loss enhances SNN prediction consistency and efficiency.} 
    \textbf{a}, Model performance over readout time, contrasting accuracy (solid line) and consistency entropy (dotted line) between models trained with fidelity (cyan) and without (yellow).
    \textbf{b}, Probability of correct predictions over readout time, using the samples from \figurename~\ref{fig:3}\textbf{b} as inputs. Solid and dashed lines indicate right and wrong classifications, respectively, by MNN models, with fidelity (cyan) and without (yellow).
    \textbf{c}, Pearson correlation coefficients comparing SNN convergence times with MNN confidence metrics. 
    The convergence time of SNNs is defined as the time when the SNN's prediction remains stable after that.
    \textbf{d}, Violin plots delineating SNN convergence times correlated with correct (red) and incorrect (black) final predictions.}
    \label{fig:4}
\end{figure}

We initially examine the progression of the prediction accuracy with increasing readout times, as shown in \figurename~\ref{fig:4}\textbf{a} (solid line).
The accuracy, assessed at the dataset level, is averaged across 100 trials. 
Although the final accuracy differences between the \textit{WithFidelity} and \textit{MeanOnly} SNN models are minimal and reflect those observed in their MNN counterparts across all three tasks (refer to Table \ref{tab:S2} for a numerical comparison), and the dynamics of their accuracy convergence reveal the influence of fidelity-entropy loss.
For simpler tasks such as MNIST, the fidelity-entropy loss yields no discernible advantage, with both models reaching high accuracy quickly. 
However, as task complexity increases, the \textit{WithFidelity} model (cyan) exhibits a faster increase in accuracy during the initial stages compared to the \textit{MeanOnly} model (yellow). 
This early advantage diminishes over time. 
While this effect is subtle in F-MNIST, it becomes more pronounced in CIFAR10, indicating that the impact of fidelity-entropy loss is amplified by task difficulty.

The observed phenomenon can be attributed to differences in prediction consistency, as shown by the dotted lines in \figurename~\ref{fig:4}\textbf{a} (dotted line). 
We measure prediction consistency through the entropy calculated from the probability of the predicted categories in 100 trials at each readout time step [Eq.~(\ref{eq:consistency_entropy}) in Methods].
Lower entropy indicates a higher likelihood of consistent predictions. 
Similar to the accuracy trends, the consistency entropy for both models diminishes with increasing readout time. 
However, the \textit{WithFidelity} SNN exhibits a more rapid decrease in entropy compared to the \textit{MeanOnly} SNN. 
This distinction becomes particularly pronounced in F-MNIST and CIFAR10, tasks with relatively lower accuracy compared to MNIST.

The observed improvements in accuracy convergence speed and prediction consistency are systematically derived from model training, not coincidence or selective reporting.
We revisit the illustrative cases from \figurename~\ref{fig:3}\textbf{b} to analyze the correct classification probabilities of the corresponding SNNs in \figurename~\ref{fig:4}\textbf{b}.
Although the spiking patterns of the neurons in the last hidden layer exhibit irregularity (Fig. \ref{fig:s1}), the results are consistent with our theoretical investigation when using the linear decoder for readout.
In the optimal case with well-optimized mean and covariance, the sample that was correctly classified by the \textit{WithFidelity} MNN shows a higher initial correct classification probability and a slightly faster convergence in \textit{WithFidelity} SNN (left panel, solid lines)
On the contrary, samples misclassified by \textit{ MeanOnly} MNN show a rapid decrease in the correct prediction probability in the corresponding SNN, revealing a misplaced confidence in incorrect predictions.
The \textit{WithFidelity} SNN, however, gradually increases the likelihood of correct decisions (dashed lines), facilitated by near-boundary mean readouts and covariance adjustments conducive to accurate inferences.
However, when optimizing only the mean or covariance (middle panel), it can be difficult to clearly distinguish the advantage of \textit{WithFidelity} SNN. 
In such cases, there may be lower probabilities of early correct predictions (when MNNs make the correct decision) or prolonged periods of struggling to avoid consistently making the wrong decision.
In the worst scenarios (bottom panel), \textit{WithFidelity} SNNs rarely achieve a correct prediction likelihood of more than fifty percent, even when MNNs make accurate predictions, also erroneously increasing confidence in incorrect predictions.

The case studies have demonstrated the predictive capability of MNN output on SNN performance. Subsequently, we investigated this association by calculating the Pearson correlation between the convergence time of the SNN predictions and the confidence measures obtained from the corresponding MNN outputs. 
In this context, the convergence time of a decision indicates the readout time point at which the SNN's choice stabilizes. 
If a prediction does not stabilize, its convergence time is identified as the final time step. As depicted in \figurename~\ref{fig:4}\textbf{c}, a negative correlation is observed between the convergence time and metrics such as DV Mean, Fidelity and Softmax, while DV Std and Entropy exhibit a positive correlation, consistent with our theoretical predictions [Eq. (\ref{eq:minimal_time})]. 
Notably, the correlation coefficients for mean-related metrics significantly surpass those for covariance-related metrics, emphasizing the predominant impact of the mean on convergence time. Furthermore, the correlations for DV Std and Entropy are more pronounced in the model trained with fidelity-entropy loss compared to those without it, highlighting the effectiveness of fidelity-entropy loss in enhancing the covariance structure of the readouts. 
Additionally, the correlation between covariance-related metrics (DV Std and Entropy) and convergence time is stronger in CIFAR10 than in MNIST and F-MNIST, indicating that covariance plays a more significant role in influencing convergence time in challenging tasks.

In our theoretical framework, together with the observed relationships between convergence times and MNN metrics, which have shown the ability to differentiate between potential model prediction outcomes (see \figurename~\ref{fig:3}\textbf{c}), it appears that there are distinct distributions of convergence times for accurately and inaccurately classified samples.
We investigated these distinctions by comparing the convergence times for correct and incorrect predictions. 
The resulting violin plots (\figurename~\ref{fig:4}\textbf{d}) demonstrate that accurately predicted samples typically converge more quickly with a positive skew, in contrast to the wider distribution observed for misclassified instances. 
Additionally, the \textit{WithFidelity} SNN demonstrates a faster convergence time in comparison to the \textit{MeanOnly} SNN, especially evident in the challenging task CIAFR10.
Due to the variability in convergence times across samples, using a fixed predefined inference duration must be determined based on the longest convergence time of the accurately classified sample to achieve the highest overall accuracy. 
However, this approach cannot be the most energy-efficient since many samples converge sooner. 
This underscores the importance of implementing adaptive methods that leverage different convergence times to improve energy efficiency.

In summary, we have demonstrated that models trained with fidelity-entropy loss lead to marked improvements in accuracy and prediction consistency. These gains align with our theoretical insights from MNNs and are further evidenced by the observed variability in convergence times for different inputs. 
This variability highlights the imperative for an adaptive stopping policy, which offers a way to significantly boost energy efficiency and sharpen the accuracy of predictions, a conception we term the free-response paradigm. 
\subsection{Efficiency and confidence expression in SNNs through adaptively stopping inference}
Moving towards the free-response paradigm requires accurate identification of when a prediction converges, a task that comes with notable difficulties. 
Unlike theoretical models such as DDMs, where readout distributions are predefined, real-world situations make it challenging to determine from which distribution an instantaneous readout comes. 
Although it is possible to precisely determine these distributions by using extended sampling periods or ensemble methods \cite{Legenstein2014}, these approaches contradict the primary goal of the free-response paradigm, which is to conserve time and energy.

In our theory, our theoretical framework addresses the challenge by focusing on the disparity between the top two readout units, essentially the magnitude of the DV.
As this amplitude increases, it becomes less likely for fluctuations in the instantaneous readout \(\mathbf{r}\) to alter the DV's sign, prompting the implementation of a 'Difference policy'. 
According to this policy, the inference process halts once the amplitude exceeds a predefined threshold, indicating convergence.

This method is compared to various alternative strategies: Fixed, Max, Linear, and Oracle. 
The Fixed strategy follows a predetermined inference duration, stopping the inference process at a specific time regardless of convergence. 
The Max strategy stops inference when the highest readout unit surpasses a certain threshold. 
The Linear strategy entails training a linear classifier to differentiate between converged and unconverged predictions based on sorted readouts, stopping inference when the classifier's score exceeds a specific threshold. 
It is worth noting that both the Difference and Max strategies can be viewed as specific cases of the Linear strategy. 
The Oracle strategy represents an ideal scenario where inference is always terminated precisely at the moment of convergence, serving as a benchmark for evaluating the maximum efficiency achievable by any stopping policy. 
Evaluating the effectiveness of these strategies by applying them to the \textit{WithFidelity} SNN model described in the previous section allows for a comprehensive comparison of their effects on the inference process.

\begin{figure}
    \centering
    \includegraphics[width=\linewidth]{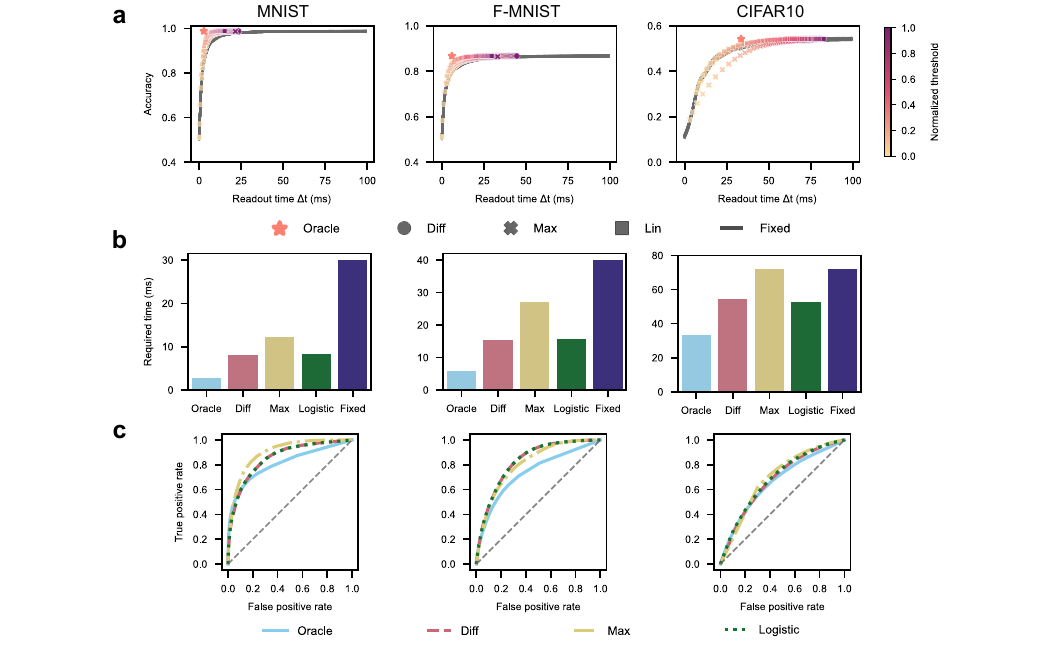}
    \caption{\textbf{Efficiency of different stopping policies in SNNs.}
    \textbf{a,} Performance comparison of various stopping policies with adjustable decision thresholds. The Oracle policy (orange star) represents an ideal scenario where inference stops at convergence. The Difference (Diff) policy (circle dot) stops inference when the gap between the top two readouts crosses a threshold. The Maximum (Max) policy halts when the highest readout exceeds its threshold, and the Linear (Lin) policy employs a linear classifier on sorted readouts, stopping when the classifier's score surpasses a predefined threshold. For comparison, the Fixed policy (gray line) represents a uniform stopping time for all samples.
    \textbf{b,} Minimum readout times required for the Diff, Max, and Lin policies to achieve accuracy within 0.5\% of Oracle performance.
    \textbf{c,} Receiver Operating Characteristic (ROC) curves that evaluate the expression of prediction confidence based on stopping times determined by policies, using thresholds from \textbf{b}.}
    \label{fig:5}
\end{figure}

As shown in Figure~\ref{fig:5}\textbf{a}, we compared the performance dynamics of various stopping policies by setting different decision thresholds (normalized for comparison).
The Oracle policy (orange star) achieves peak accuracy with the shortest time requirement, setting an ideal benchmark.
For high-accuracy tasks like MNIST and F-MNIST, Difference, Max, and Linear policies outshine the Fixed approach by delivering quicker and more accurate outcomes. 
The Linear policy demonstrates a slight advantage over the others in terms of accuracy improvements when the constraint of limited readout time is considered (averaged across the test set). 
As decision thresholds are raised, these policies move closer to achieving accuracy levels similar to an Oracle, suggesting a trade-off between accuracy and time usage. 
In contrast, when faced with challenging tasks like CIFAR10, the Max policy falls behind, particularly at lower thresholds where the emphasis shifts more towards speed. On the other hand, the Difference and Linear policies maintain their superiority, delivering enhanced accuracy in a more efficient manner. 
This underscores the adaptability of the Difference and Linear policies across varying levels of complexity, highlighting their potential to enhance accuracy while ensuring swift convergence. 
Furthermore, although the Linear policy slightly outperforms our Difference policy when the threshold configuration prioritizes speed, it necessitates the training of a suitable classifier beforehand. 
This makes the Difference policy superior, as only a proper decision threshold is required.

To evaluate the comparative effectiveness of different stopping strategies, we investigated the minimum time taken by each strategy to approximate the accuracy achieved by the Oracle strategy. 
A tolerance threshold (0.005) is established to define when a strategy reaches Oracle-level accuracy. 
This threshold is chosen to strike a balance between the importance of improvement and the amount of time spent. 
Figure~\ref{fig:5}\textbf{b} illustrates that both the Difference and Linear policies outperform the Fixed policy in terms of the time required to achieve the desired accuracy in all tasks considered. 
Although the Linear strategy shows slightly better efficiency compared to the Difference strategy, this advantage is practically negligible. 
Conversely, the Max strategy does reduce the time needed compared to the Fixed strategy for simpler tasks, but it still requires more time to converge than either the Difference or Linear strategies. 
Importantly, the Max strategy does not provide any time-saving benefit in more challenging scenarios, where the upper bound of model performance is limited. 
This illustrates the effectiveness of our Difference policy, which remains quite strong across different tasks, despite the gap when compared to the Oracle policy.
It is important to mention that as stopping policies operate during the inference phase, we also confirmed the effectiveness of our Difference policy in SNN models trained using various techniques~\cite{Hao2023, li2022converting} (as outlined in Table \ref{tab:S3}).

In addition to facilitating time efficiency, the implementation of stopping policies offers the advantage of using stopping time as an indicator of prediction accuracy. 
To evaluate the effectiveness of stopping time as a predictive quality metric, we used ROC analysis.
Figure~\ref{fig:5}\textbf{c} reveals that the AUROC for all policies in the three tasks exceeds 0.5, indicating their predictive validity. 
This is particularly evident in CIFAR10, where the number of correct and incorrect predictions is nearly even. 
It is important to note that in simple tasks like MNIST and F-MNIST, the Oracle policy, which optimally stops inference, shows a lower AUROC compared to other policies. 
This discrepancy is due to the limited time resolution of the simulation and the relatively low task complexity, which can lead both correct and incorrect samples to potentially converge within the same time steps, making their differentiation more challenging. 
In contrast, our previous findings suggest that the convergence time is mainly influenced by the mean disparity, and increasing the decision threshold in a manually designed stopping policy allows for a more effective distinction by requiring an extended time to achieve it.
Despite variations in the average readout time required by each policy to achieve comparable accuracy levels, the AUROC values for discerning the quality of the model's predictions remain consistently high across policies. 
This shows a direct correlation between shorter stopping times and a higher probability of correct predictions and longer stopping times with higher probabilities of incorrect outcomes.
\section{Disscussion}
In this work, we have formulated a comprehensive theory that delineates the decision-making mechanisms within SNNs. 
This theory dissects the decision-making process into two fundamental inquiries: identifying the optimal choice based on the present evidence and determining the confidence level of that choice relative to the available alternatives. 
Our contributions notably refine SNN applications across two critical phases:
In the training phase, we introduce a novel training objective that encourages SNNs not only to discern the correct decisions, but also to cultivate a robust confidence measure derived from historical training performance. 
During the inference stage, our results analyze traditional SNN inference methods, noting the lack of a dynamic stopping mechanism similar to decision thresholds in DDMs. 
We highlight the importance of the gap in the readout between the top two outputs as a critical factor. 
Through the integration of adaptive stopping policies, our SNNs shift from a predetermined duration to a more flexible response paradigm, resulting in notable time savings. 
Importantly, we identify a reverse relationship between convergence speed and prediction confidence, echoing findings from studies in neurobiology~\cite{Gold2007, Kepecs2008, Kiani2014}. 

Confidence indicates the ability of a neural system (whether biological or artificial) to assess uncertainty in decision-making, which is essential to influence future actions.
Predominant neuroscience theories posit that confidence is equal to the likelihood that a decision is correct~\cite{Kepecs2012, Mamassian2016, Pouget2016}.
However, our theory proposes a nuanced perspective, suggesting that confidence is determined by the disparity between the first and second best choices. 
This distinction becomes particularly pronounced in multi-choice scenarios, where traditional and our approaches diverge, especially under conditions of low confidence. 
The conventional view interprets confidence as the comparative likelihood of one choice over all others, whereas our theory emphasizes the direct comparison between the top two contenders. 
In situations where the top two options are nearly equally probable, diverging significantly from the remainder, our framework and traditional models may infer contrasting levels of confidence. 
This observation aligns with the findings of Li et al.~\cite{Li2020}  in a three-option visual categorization task, underscoring the nuanced nature of confidence.

Additionally, our experimental findings suggest that the ability to express confidence is greatly influenced by the accuracy of the model. This means that uncertainty in decision-making can be categorized into two types: objective uncertainty and subjective uncertainty. 
Objective uncertainty refers to the doubt regarding whether the decision made by a neural system is correct or incorrect, while subjective uncertainty pertains to the system's confidence in its decision. 
These two forms of uncertainty may align if the neural system has fully grasped the true posterior probability of the selected option, a feat that is almost unattainable in practical scenarios~\cite{Beck2012}. 
The absence of a definitive posterior probability for decisions also poses a significant challenge in improving the confidence calibration of models in artificial intelligence~\cite{Gawlikowski2023}.
Hence, confidence is solely based on the subjective belief of a neural system, and it is essential to evaluate both the neural system's performance and its confidence level before endorsing its decision.

Expanding on the fundamental importance of confidence in decision-making, we now focus on the stopping policy of the inference process, a concept with a rich historical background.
The notion of stopping inference dynamically traces its roots to sequential analysis, a mathematical framework for adaptive decision rules developed during World War II~\cite{Wald1973}.
This framework has evolved into an integral component of statistical decision theory~\cite{Poor1994}, underscoring the potential of adaptive decisions to enhance readout schemes by facilitating real-time monitoring and adjustments over considerably shorter time scales than those required for high-fidelity readouts\cite{D’Anjou2016}.
Despite its established significance, the application of adaptive decision-making within the realm of neuromorphic computing has not been explored much~\cite{li2023unleashing, li2024seenn}.
In this regard, our research aims to connect this divide by utilizing the extensive knowledge of sequential analysis and adaptive decision theories to explore the spatio-temporal patterns of SNNs, an area with great potential for new ideas and advancements.

Although our findings offer valuable insights, we recognize certain limitations that merit attention. 
Our demonstrations were based on the moment neural network (MNN) framework~\cite{Feng2006, Lu2010, Qi2022}, which, despite its utility, faces scalability challenges that hinder the training of more sophisticated networks for complex tasks. 
Furthermore, the fidelity-entropy loss that we introduced may not seamlessly integrate into the training of SNNs via the conventional ANN-SNN conversion method due to the inability of ANNs to mimic the covariance propagation characteristic of MNNs.
Future efforts could explore surrogate gradient methods to incorporate fidelity-entropy loss in SNN training, allowing for the approximation of mean and covariance through sampling techniques.
Additionally, our use of a fixed decision threshold in stopping policy represents an area ripe for innovation. 
An adaptive threshold, modulated in response to the accuracy of recent decisions, could more accurately reflect the dynamic decision-making processes observed in biological systems, as evidenced by animal behavior studies~\cite{Vickers1982, Deco2009, Gupta2024}.
This self-adjusting mechanism has the potential to allow SNNs to automatically determine the best thresholds, overcoming the difficulties of manual calibration in unpredictable input situations and when deploying SNNs on various hardware platforms.

In summary, this study establishes a new connection between cognitive neuroscience and neuromorphic computing in the context of decision-making. By highlighting the impact of signal (mean) and noise (covariance) on decision processes, we argue that simply averaging SNN outputs is insufficient for fully capturing the diverse spatio-temporal dynamics. 
Our results support a more nuanced approach for utilizing SNN capabilities, offering a more comprehensive exploration of the potential of brain-inspired computing.

\section{Methods}
\subsection{Leaky integrate-and-fire neuron model}
We used the leaky integrate-and-fire (LIF) spiking neuron model to describe the membrane potential dynamics of neurons:
\begin{equation}
    \dfrac{dV_i}{dt}= -LV_i(t) + I_i(t),
    \label{eq:LIF}
    \end{equation} 
where the sub-threshold membrane potential $V_i(t)$ of a neuron $i$ is driven by the total synaptic current $I_i(t)$ and $L=0.05$ \si{\per\milli\second} is the leak conductance. 
When the membrane potential $V_i(t)$ exceeds a threshold $V_{\rm th}=20$ \si{\milli\volt} a spike is emitted, as represented by a Dirac delta function. Afterward, the membrane potential $V_i(t)$ is reset to the resting potential $V_{\rm res}=0$ mV, followed by a refractory period $T_{\rm ref}=5$ ms. The synaptic current takes the form
\begin{equation}
    I_i(t)= \sum_{j}w_{ij}S_j(t)+I_i^{\rm ext}(t),
    \label{eq:current}
\end{equation}
where $S_j(t)=\sum_k \delta(t-t^k_j)$ represents the spike train generated by presynaptic neurons.

A final output $\mathbf{y}$ is readout from the spike count $\mathbf{n}(\Delta t)$ of a population of spiking neurons over a duration time window $\Delta t$ as follows:
\begin{equation}
y_i(\Delta t) = 
\frac{1}{\Delta t}\sum_{j} w_{ij}n_{j}(\Delta t) + \beta_i, 
\label{eq:snn_readout}
\end{equation}
where $w_{ij}$ and $\beta_i$ are the weights and biases of the readout, respectively. 
A key characteristic of the readout is that its variance decreases as the time window $\Delta t$ increases.
\subsection{Moment embedding for the leaky integrate-and-fire neuron model}
The moment embedding approach~\cite{Feng2006, Lu2010, Qi2022} begins with mapping the fluctuating activity of spiking neurons to their respective first and second order moments:
\begin{equation}
\mu_i = \lim_{\Delta t\to\infty} \dfrac{\mathbb{E}[n_i(\Delta t)]}{\Delta t}, 
\label{eq:def_mean}
\end{equation}
and
\begin{equation}
\Sigma_{ij} = \lim_{\Delta t\to\infty} \dfrac{{\rm Cov}[n_i(\Delta t),n_j(\Delta t)]}{\Delta t}, 
\label{eq:def_cov}
\end{equation}
where $n_i(\Delta t)$ is the spike count of neuron $i$ over a time window $\Delta t$. 
In practice, the limit of $\Delta t\to\infty$ is interpreted as a sufficiently large time window relative to the timescale of neural fluctuations. 
We refer to the moments $\mu_i$ and $\Sigma_{ij}$ as the mean firing rate and the firing covariability in the context of MNN, respectively. 

For the LIF neuron model [Eq.(\ref{eq:LIF})], the statistical moments of the synaptic current are equal to~\cite{Feng2006, Lu2010}
\begin{numcases}{}
\bar{\mu}_i = \textstyle\sum_kw_{ik}\mu_k + \hat{\mu}_i^{\rm ext},\label{eq:sum_mean}
\\
\bar{\Sigma}_{ij}=\textstyle\sum_{kl} w_{ik}C_{kl}w_{jl} + \hat{C}_{ij}^{\rm ext},\label{eq:sum_cov}
\end{numcases}
where $w_{ik}$ is the synaptic weight and $\hat{\mu}_i^{\rm ext}$ and $\hat{\Sigma}_{ij}^{\rm ext}$ are the mean and covariance of an external current, respectively.  
Note that from Eq. (\ref{eq:sum_cov}), it becomes evident that the synaptic current is correlated even if the presynaptic spike trains are not. 
Next, the first- and second-order moments of the synaptic current are mapped to those of the spiking activity of the postsynaptic neurons. 
For the LIF neuron model, this mapping can be obtained in closed form through a mathematical technique known as the diffusion approximation~\cite{Feng2006, Lu2010} as
\begin{numcases}{}
\hat{\mu}_i = \phi_\mu(\bar{\mu}_i,\bar{\sigma}_i),\label{eq:ma_mu}\\
\hat{\sigma}_i = \phi_\sigma(\bar{\mu}_i,\bar{\sigma}_i),\label{eq:ma_sig}\\
\hat{\rho}_{ij} = \chi(\bar{\mu}_i,\bar{\sigma}_i)\chi(\bar{\mu}_j,\bar{\sigma}_j)\bar{\rho}_{ij},\label{eq:ma_chi}
\end{numcases}
where the correlation coefficient \(\hat{\rho}_{ij}\) is related to the covariance as \(\hat{\Sigma}_{ij}=\hat{\sigma}_i\hat{\sigma}_j\rho_{ij}\). The mapping given by Eq. (\ref{eq:ma_mu})-(\ref{eq:ma_chi}) is called moment activation, which is differentiable so that gradient-based learning algorithms can be implemented and the learning framework is known as the moment neural network (MNN)~\cite{Qi2022}.

\subsection{Model setup for MNN training}
We developed the MNN framework using Pytorch to leverage backpropagation for optimization. 
MNNs were trained on three datasets: MNIST~\cite{Lecun1998},  Fashion-MNIST (F-MNIST)~\cite{xiao2017fashion} and CIFAR10~\cite{krizhevsky2009learning}. 
For MNIST and F-MNIST, the MNN architecture included one hidden layer, while for CIFAR10, it comprised three hidden layers, each containing 1000 LIF neurons in a feedforward configuration.

The MNIST dataset features 60,000 training and 10,000 testing images of handwritten digits, whereas Fashion-MNIST provides 60,000 training and 10,000 validation images of grayscale clothing items. 
CIFAR10 offers a collection of natural images, with 50,000 for training and 10,000 for validation. 
The input dimension for the MNNs corresponds to the image size of each dataset (784 for MNIST and F-MNIST, 3072 for CIFAR10).

Input images \(\mathbf{x}\)  were encoded using a Poisson-rate scheme, with mean \(\mathbf{\mu}\) and variance \(\mathbf{\sigma}^2\) mirroring pixel intensities. 
Off-diagonal correlation coefficients were set to zero. 
Given that all datasets encompass 10 categories, a linear readout function translated the last hidden layer outputs into a ten-dimensional space for classification using the operations in Eq.~(\ref{eq:sum_mean})-(\ref{eq:sum_cov}), which were the outputs of the MNNs.

Using the MNN output (mean \(\hat{\mathbf{\mu}}\) and covariance \(\hat{\Sigma}\)), we applied the loss function described by Eq.~\ref{eq: fidelity_loss}, comparing models trained with (\textit{WithFidelity}) and without (\textit{MeanOnly}) the fidelity-entropy loss.
Crucially, the fidelity-entropy loss calculation involved sorting \(\hat{\mathbf{\mu}}\) and \(\hat{\Sigma}\))in descending order based on \(\hat{\mathbf{\mu}}\).
The primary weight \(w_1\) was set to 0.8, emphasizing the fidelity entropy between the top two entries, while the remaining pairs received a weight of \(w_j = 0.2 / 9, \forall j > 1\).
For all experiments, the readout time \(\Delta t\) in the loss function was standardized to 1.

Each dataset underwent 5 experimental runs under both conditions (\textit{WithFidelity} and \textit{MeanOnly}), with each experiment consisting of 30 epochs and a batch size of 50. 
Random crop augmentation was the only technique used during training in which images were first padded with two pixels on all sides, then randomly cropped back to their original size. 
Model parameters were optimized using the AdamW optimizer, adhering to the default settings (a learning rate of 0.001 and a weight decay of 0.01).

\subsection{Data analysis in MNN models}
Upon obtaining the model's readouts \(\hat{\mathbf{\mu}}\) and \(\hat{\Sigma}\), we determined its predictions by identifying the largest entry of \(\hat{\mathbf{\mu}}\).
This approach allowed us to evaluate the prediction accuracy of both \textit{WithFidelity} and \textit{MeanOnly} models on the test sets, with results depicted during training in Fig.~\ref{fig:3}\textbf{a} and summarized after training in Table \ref{tab:S1}. 

For in-depth analysis, we utilized models trained for 30 epochs, as shown in Fig.~\ref{fig:3}\textbf{c} and \textbf{d}.
We derived metrics including DV Mean, DV Std, and Fidelity directly from Eq.~\ref{eq: fidelity}, based on the top two entries of \(\hat{\mathbf{\mu}}\) and \(\hat{\Sigma}\)) in terms of \(\hat{\mathbf{\mu}}\), and processed test set samples to compute these values. 
Entropy was calculated from \(\hat{\Sigma}\)), using the formula:
\begin{equation}
  H = \frac{1}{2} \ln ((2\pi e)^n |\Sigma|),
  \label{eq:gaussian_entropy}
\end{equation}
where \(n\) is the dimension of \(\Sigma\).
The Softmax score was determined by applying the softmax function to \(\hat{\mathbf{\mu}}\) and selecting the largest entry (Eq.~\ref{eq:softmax_func}).
Subsequently, we categorized the test set samples on the basis of the results of the models' predictions.
We calculated their AUROC scores using Scipy's built-in functions to quantify the efficacy of these metrics in guiding decision acceptance.

To evaluate the potential improvement of any input using fidelity-entropy loss, we analyzed the proportion of test set instances in which performance metrics such as DV Mean, Fidelity, and Softmax were superior in models that include fidelity (\(WithFidelity\)) as opposed to those without it (\textit{MeanOnly}), specifically for correctly classified instances. 
Conversely, these metrics are expected to be lower for misclassified instances.
The same comparison was conducted for DV Std and Entropy, albeit with opposite implications because of their inherent characteristics.

\subsection{Experiments setup for SNN simulation}
The trained MNN parameters were used to reconstruct the SNN models simply by replacing the moment activation layer in MNN with LIF neurons, as explained in our previous study~\cite{Qi2022}. 
The trained MNN models for the case studies shown in Figure \ref{fig:3}\textbf{b} were selected from the \textit{WithFidelity} and \textit{MeanOnly} conditions to generate the respective SNNs.
These SNNs are now referred to as the \textit{WithFidelity} SNN and \textit{MeanOnly} SNN, respectively.

The spike trains used as input for these SNNs were created using a Poisson process with rates that mirror the pixel intensities of the images in the test datasets. 
This method is consistent with the original training principles of the MNNs. 
Evaluation of each instance was carried out within a timeframe of 100 \si{\milli\second}, with a time resolution of \(\delta t = 1\) \si{\milli\second}, resulting in 100 instantaneous readout values for each instance.

The starting membrane potentials of LIF neurons were randomly assigned within the range of 0 \si{\milli\volt} and their respective firing thresholds. 
Throughout the simulation, the membrane potentials of LIF neurons in the SNNs did not reset to zero after processing each sample in a test set. 
This methodology was consistently applied to all tasks and the simulation was replicated 100 times with different initial membrane potentials.
The resulting instantaneous readout values \(\mathbf{r}\) from these simulations were then examined for additional analysis.

\subsection{Data analysis in SNN simulations}
To analyze the readouts from the SNN simulations, we first aggregated the instantaneous readout values \(\mathbf{r}(t)\) over time to calculate the cumulative readout \(\mathbf{d}(t)\).
By identifying the index of the largest entry in \(\mathbf{d}(t)\) as the model prediction \(y_{\rm pred}\), and comparing it with the ground truth at each time point \(t\), we thus obtained the accuracy curves for the model.
Prediction consistency at time\(t\) was quantified using the entropy measure \(H(t)\):
\begin{equation}
  H(t) = -\sum_{k}P^{(t)}(y_{\rm pred}=k)\log P^{(t)}(y_{\rm pred}=k),
  \label{eq:consistency_entropy}
\end{equation}
where, \(P^{(t)}(y_{\rm pred}=k)\) denotes the probability of the model predicting category \(k\) at time \(t\), determined by counting the occurrences of each predicted category in 100 trials. 
We computed the mean values of both the accuracy and prediction consistency entropy curves for all samples in the test set, including confidence intervals to provide a thorough analysis.

For the case studies depicted in Fig.~\ref{fig:4}\textbf{b}, we reused the samples highlighted in Fig.~\ref{fig:3}\textbf{b}, assessing the likelihood of correct predictions in each scenario.

The convergence time from these simulations was defined as the earliest moment \(t^*\) when the prediction of a model remained constant until the end of the simulation period ( \(t=100\) \si{\milli\second}).
This procedure yielded a convergence time for each sample and trial. 
Subsequently, we examined the Pearson correlation between these convergence times and the five metrics previously derived from the corresponding MNN models for the input samples~(Fig. \ref{fig:4}\textbf{c}).
Furthermore, by categorizing convergence times based on the outcomes of the corresponding predictions (right or wrong), we visualized the distribution of convergence times~(Fig. \ref{fig:4}\textbf{d}), offering insights into the temporal dynamics of prediction stabilization in SNN simulations.

\subsection{The implementation of stopping policies}
Utilizing the cumulative readout \(\mathbf{d}(t)\)  from the \textit{WithFidelity} SNN, the Difference policy terminates inference when \(d_i -d_j > \Theta\) at time \(t\), where \(d_i\) and \(d_j\) represent the largest and second-largest entries, respectively.
 The Max policy ceases inference if \(d_i < \Theta\) at time \(t\), focusing solely on the largest entry. 
For the Linear policy, after sorting \(\mathbf{d}(t)\) at each time \(t\) to have a consistent rank,  to ensure a consistent ranking, we classified the readouts into converged and non-converged based on the convergence times previously determined. 
A logistic classifier was trained on these labels using the cumulative readout, and inference was halted when its output score exceeded the threshold \(\Theta\). 
The Oracle policy, leveraging the actual convergence time, serves as the benchmark for comparing the efficacy of these policies.

We explored a wide range of decision thresholds for each policy to span the spectrum from low to high accuracy. 
For the Difference and Max policies,\(\Theta\) varied between \([10, 100]\), whereas for the Linear policy, it ranged from \([-1.5, 10]\).
Thresholds were normalized for consistent color coding in Fig. \ref{fig:5}\textbf{a}.
Based on the speed-accuracy curves, we analyzed the mean time required by each policy to match Oracle accuracy, as depicted in Fig. \ref{fig:5}\textbf{b}, and conducted AUROC analysis on their stopping times to evaluate confidence expression (Fig. \ref{fig:5}\textbf{c}).

In additional experiments (Table \ref{tab:S3}), we used the ANN-SNN conversion methodology introduced by Hao et al.~\cite{Hao2023} and Li et al.~\cite{li2022converting}, available at \href{https://github.com/hzc1208/ANN2SNN\_SRP}{https://github.com/hzc1208/ANN2SNN\_SRP} and \href{https://github.com/yhhhli/SNN\_Calibration}{https://github.com/yhhhli/SNN\_Calibration}. 
When working with the MNIST dataset, we replicated the experimental configuration as previously outlined, training ANNs for 30 epochs. 
For CIFAR100~\cite{krizhevsky2009learning}, we followed the parameters set by Hao et al.~\cite{Hao2023}. 
In the case of these two datasets, the conversion of ANNs to SNNs was carried out using the approach proposed by Hao et al., known as the residual membrane potential (SRP) method. 
When dealing with the ImageNet dataset~\cite{Deng2009}, we employed pretrained models from Li et al.~\cite{li2022converting}, and implemented their light calibration pipeline (LCP).

Crucially, in these experiments, the inputs remained static, matching the pixel intensities of the images, eliminating the variability introduced by noisy inputs. 
For MNIST and CIFAR100, simulations ran for 100 steps, while ImageNet required 256 steps. 
After simulation, we applied the Difference policy to these converted SNNs, assessing the time to achieve 95\% of the final accuracy to demonstrate the applicability of our theory across different training methodology and network architecture.

\section*{Data availability}
The datasets used in this study are available in the following databases.
\begin{itemize}
    \item MNIST: \href{https://git-disl.github.io/GTDLBench/datasets/mnist\_datasets/}{https://git-disl.github.io/GTDLBench/datasets/mnist\_datasets/}
    \item Fashion-MNIST: \href{https://github.com/zalandoresearch/fashion-mnist}{https://github.com/zalandoresearch/fashion-mnist}
    \item CIFAR10 and CIFAR100: \href{https://www.cs.toronto.edu/~kriz/cifar.html}{https://www.cs.toronto.edu/~kriz/cifar.html}
    \item ImageNet: \href{https://www.image-net.org/download.php}{https://www.image-net.org/download.php}
\end{itemize}

\section*{Code Availability}
The code used in this article is available at \href{https://github.com/BrainsoupFactory/moment-neural-network}{https://github.com/BrainsoupFactory/moment-neural-network} and \href{https://github.com/ZhichaoZhu/moment-neural-network}{https://github.com/ZhichaoZhu/moment-neural-network}.

\section*{Acknowledgments}
This work is jointly supported by Shanghai Municipal Science and Technology Major Project (No.2018SHZDZX01), the National Natural Science Foundation of China (No. 62306078, No. 62072111), ZJ Lab and Shanghai Center for Brain Science and Brain-Inspired Technology.

\section*{Author Contributions}
Conceptualization: Z.Z., Y.Q. and J.F.; 
Methodology: Z.Z., Y.Q., W.L., and J.F.;
Investigation: Z.Z.;
Software: Z.Z. and Y.Q.;
Visualization: Z.Z.;
Resources: Y.Q, L.C., Z.G. and J.F.;
Writing — original draft: Z.Z.;
Writing — review \& editing: Z.Z., Y.Q., W.L., Z.G. and J.F.;
Supervision: J.F.

\section*{Competing interests}
The authors declare that they have no competing interests.

\section*{Materials \& Correspondence}
Correspondence and materials requests should be sent to J.F.
\newpage
\bibliography{reference}
\bibliographystyle{naturemag}

\newpage
\begin{flushleft}
{\Large
\textbf\newline{{\bf Supplementary Information}}
}
\end{flushleft}
\setcounter{section}{0}
\renewcommand{\thesection}{S\arabic{section}}
\setcounter{equation}{0}
\renewcommand{\theequation}{S\arabic{equation}}
\setcounter{figure}{0}
\renewcommand{\thefigure}{S\arabic{figure}}
\setcounter{table}{0}
\renewcommand{\thetable}{S\arabic{table}}

\begin{table}[h!]
\begin{center}
\caption{The accuracy of MNNs trained with and without fidelity-entropy loss}
\label{tab:S1}
\smallskip
\begin{tabular}{cccc}
\toprule
 & MNIST & F-MNIST & CIFAR10 \\ \hline
WithFidelity & 98.75\% \(\pm\) 0.08\% & 86.92\% \(\pm\) 0.28\% & 55.62\% \(\pm\) 0.73\% \\ \midrule
MeanOnly & 98.72\% \(\pm\) 0.11\% & 87.01\% \(\pm\) 0.39\% & 54.93\% \(\pm\) 1.61\% \\ \bottomrule
\end{tabular}
\end{center}
\noindent
\smallskip
* The accuracy is assessed by models that were trained for 30 epochs and is indicated by a 95\% confidence interval.
\end{table}

\begin{table}[h!]
\caption{Comparision of the reconstructed SNNs' accuracy at the final step}
\label{tab:S2}
\begin{center}
    \begin{tabular}{cccc}
\hline
\textbf{Dataset}& \textbf{Loss function} & \textbf{ACC at final step} & \textbf{ACC of the corresponding MNN} \\ \toprule
\multirow{2}{*}{MNIST} & WithFidelity & 98.76\% &  98.80\%\\ \cline{2-4} 
                  & MeanOnly & 98.64\% & 98.67\% \\ \hline
\multirow{2}{*}{F-MNIST} & WithFidelity & 86.72\% & 86.79\% \\ \cline{2-4} 
                  & MeanOnly & 86.04\% & 86.19\% \\ \hline
\multirow{2}{*}{CIFAR10} & WithFidelity & 54.22\% & 56.79\% \\ \cline{2-4}
                  & MeanOnly & 52.77\% & 55.7\% \\ \bottomrule
\end{tabular}
\end{center}
\noindent
\smallskip
* The accuracy of the reconstructed SNNs is evaluated by averaging the results of 100 trials.
\end{table}

\begin{table}[h!]
    \begin{center}
        \centering
        \caption{Comparison of average steps required with or without the stopping policy}
    \label{tab:S3}
    \resizebox{\textwidth}{!}{
    \begin{tabular}{ccccccc}
    \toprule
    \textbf{Dataset} & \textbf{Method} & \textbf{Arch} & \textbf{ACC at final step} & \multicolumn{1}{c}{\textbf{Average step}} &\multicolumn{2}{c}{\textbf{Average steps for 95\% ACC}} \\
     &  & & & (Oracle policy) & Fixed policy & Difference policy \\
    \midrule
    MNIST & SRP~\cite{Hao2023}\textsuperscript{*} & MLP & 98.95\%(T=100) & 4 & 4 & 2.89 \\
    CIFAR100 & SRP~\cite{Hao2023}\textsuperscript{*} & ResNet20 & 66.26\%(T=100) & 9 &11 & 10.65 \\
    ImageNet & LCP~\cite{li2022converting}\textsuperscript{**} & ResNet34 & 71.69\%(T=256) & 117.45 & 189 & 153.26 \\
    ImageNet & LCP\cite{li2022converting}\textsuperscript{**} & VGG16BN & 50.75\%(T=256) & 147.21 & 253 & 208.75 \\
    \bottomrule
    \end{tabular}
    }
    \end{center}
    \noindent
    \smallskip
    * Training from scratch.
    
    \smallskip
    ** Using the pre-trained models.
    \end{table}

\begin{figure}[h!]
    \begin{center}
        \includegraphics[width=\linewidth]{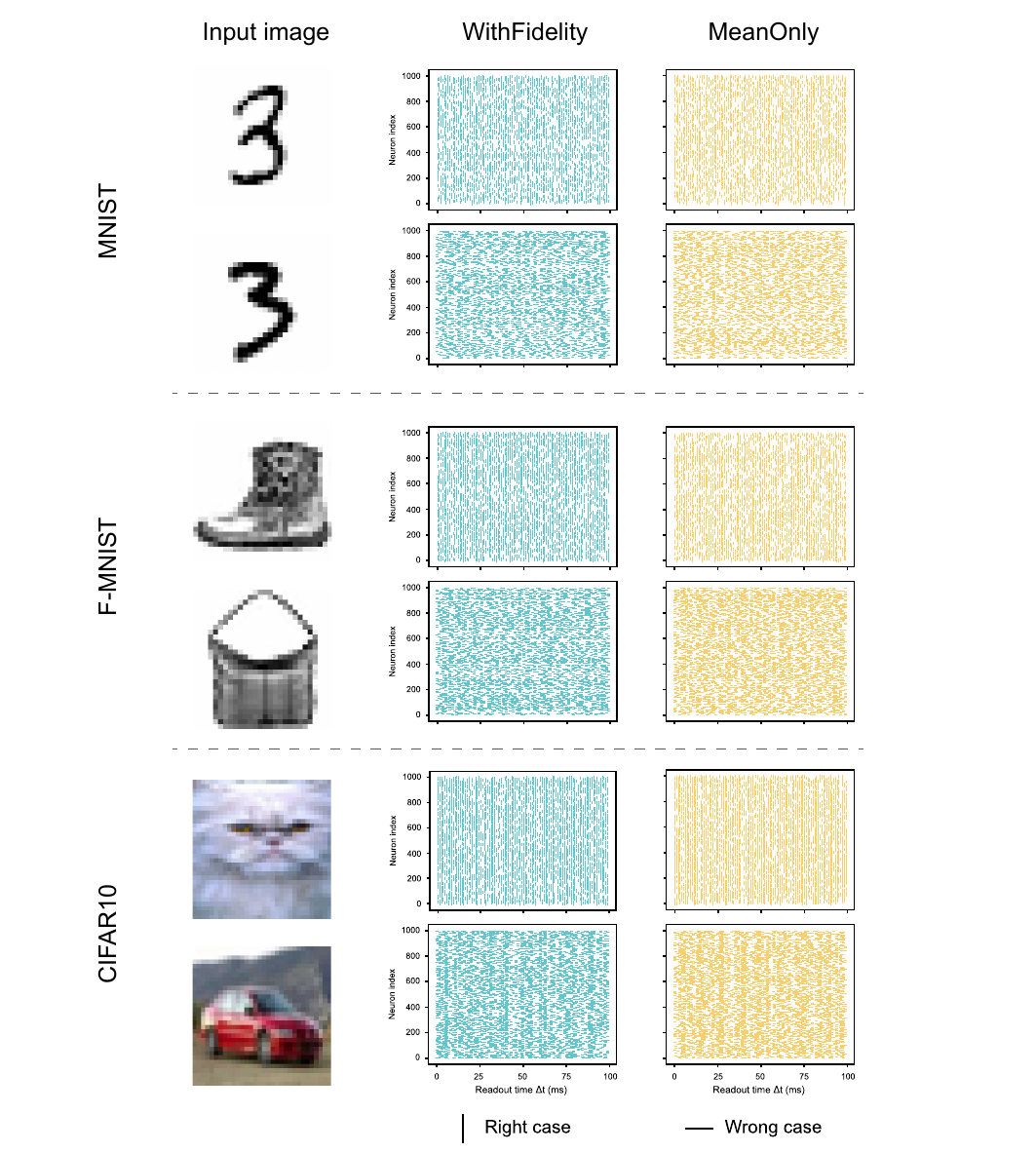}
    \end{center}
    \caption{\textbf{Raster plots illustrating the spiking behavior of SNNs in the case studies.} The input images correspond to \figurename \ref{fig:3}\textbf{b} and \figurename \ref{fig:4}\textbf{b}. The spiking activity of the final hidden neurons in the \textit{WithFidelity} SNN and the \textit{MeanOnly} SNN is depicted. Each row represents a different scenario: the top row shows a sample correctly classified by both the \textit{WithFidelity} SNN and the \textit{MeanOnly} SNN (marked with vertical lines in the raster plots), while the bottom row displays a sample misclassified by the SNNs (marked with horizontal lines in the raster plots).}
    \label{fig:s1}
\end{figure}
\end{document}